\documentclass[final]{cvpr}

\usepackage[toc, page]{appendix}

\usepackage{times}
\usepackage{epsfig}
\usepackage{graphicx}
\usepackage{amsmath}
\usepackage{amssymb}
\usepackage{nicefrac}

\usepackage{enumitem}
\usepackage{subfigure}

\usepackage{multirow}
\usepackage{tabularx}
\usepackage{color}

\usepackage[pagebackref=true,breaklinks=true,colorlinks,bookmarks=false]{hyperref}

\begin{document}

%%%%%%%%% TITLE
\title{Extreme Value Preserving Networks}

\author{Mingjie Sun\\
Carnegie Mellon University\\
{\tt\small mingjies@cs.cmu.edu}
% For a paper whose authors are all at the same institution,
% omit the following lines up until the closing ``}''.
% Additional authors and addresses can be added with ``\and'',
% just like the second author.
% To save space, use either the email address or home page, not both
\and
Jianguo Li\\
Ant Group\\
{\tt\small jglee@outlook.com}
\and
Changshui Zhang\\
Tsinghua University\\
{\tt\small zcs@mail.tsinghua.edu.cn}
}

\maketitle

%%%%%%%%% ABSTRACT
\begin{abstract}
Recent evidence shows that convolutional neural networks (CNNs) are biased towards textures so that CNNs are non-robust to adversarial perturbations over textures, while traditional robust visual features like SIFT (scale-invariant feature transforms) are designed to be robust across a substantial range of affine distortion, addition of noise, etc with the mimic of human perception nature.
This paper aims to leverage good properties of SIFT to renovate CNN architectures towards better accuracy and robustness. We borrow the scale-space extreme value idea from SIFT, and propose extreme value preserving networks (EVPNets). 
Experiments demonstrate that EVPNets can achieve similar or better accuracy than conventional CNNs, while achieving much better robustness on a set of adversarial attacks (FGSM,PGD,etc) even without adversarial training.
\end{abstract}

%%%%%%%%% BODY TEXT
\section{Introduction}

State-of-the-art CNNs are challenged by their robustness, especially vulnerability to adversarial attacks based on small, human-imperceptible modifications of the input~\cite{szegedy2013intriguing,ian2015explain}. \cite{su2018is} thoroughly study the robustness of 18 well-known ImageNet models using multiple metrics, and reveals that adversarial examples are widely existent. Many methods are proposed to improve network robustness, which can be roughly categorized into three perspectives: (1) modifying input or intermediate features by transformation \cite{guo2018input,menet}, denoising \cite{liao2018guided,jia2019comdefend}, generative models~\cite{samangouei2018defense,song2018pixel}; (2) modifying training by changing loss functions \cite{wong2017provable,elsayed2018large,zhang2019theoretically}, network distillation \cite{papernot2016distillation}, or adversarial training \cite{ian2015explain,tramer2017ensemble}  (3) designing robust network architectures~\cite{xie2018denoising, svoboda2019peernet,nayebi2017biologically} and possible combinations of these basic categories. For more details of current status, please refer to a recent survey \cite{akhtar2018threat}.

Although it is known that adversarial examples are widely existent \cite{su2018is}, some fundamental questions are still far from being well studied  like what causes it, and how the factor impacts the performance, etc. One of the interesting findings in \cite{su2018is} is that model architecture is a more critical factor to network  robustness than model size (e.g. number of layers).
Some recent works start to explore much deeper nature. For instance, both  \cite{geirhos2018imagenet,baker2018deep} show that CNNs are trained to be strongly biased towards textures so that CNNs do not distinguish object contours from other local or even noise edges, thus perform poorly on shape dominating object instances.
On the contrary, there are no statistical difference for human behaviors on both texture rich objects and global shape dominating objects in psychophysical trials.
\cite{ilyas2019adversarial} further analyze and show that deep convolutional features can be categorized into robust and non-robust features, while non-robust features may even account for good generalization. However, non-robust features are not expected to have good model interpretability.
It is thus an interesting topic to disentangle robust and non-robust features with certain kinds of \textit{human priors} in the network designing or training process.
\begin{figure*}[]
\centering
    \includegraphics[width=0.90\linewidth]{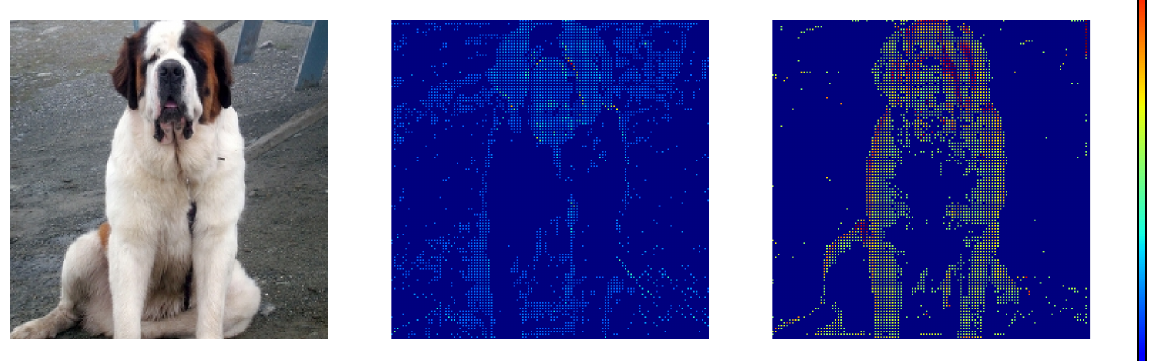}
    \vspace{1.5ex}
    \caption{Left: input image; Middle: response after ReLU of 1st conv-layer in ResNet-50; Right: response of 1st EVPConv-layer in EVPNet-50 (by replacing all $k\times k $, $k>1$, conv-layers in ResNet-50).
    Multi-channel feature maps are merged into one channel with per-pixel max-out operations for easy illustration. Clearly, ResNet-50 has more noise responses, while EVPNet-50 gives more responses on object boundary.
    }
\label{fig:cmp}
% \vspace{-3.5ex}
\end{figure*}

In fact, human priors have been extensively used in handcraft designed robust visual features like SIFT \cite{lowe2004sift}. SIFT detects scale-space \cite{lindeberg1994scale} extrema from input images,
and selects stable extrema to build robust descriptors with refined location and orientation, which achieves great success for many matching and recognition based vision tasks before CNN being reborn in 2012 \cite{alexnet12}.
The scale-space extrema are efficiently implemented by using a difference-of-Gaussian (DoG) function to search over all scales and image locations, while the DoG operator is believed to biologically mimic the neural processing in the retina of the eye \cite{young1987dog}.
Unfortunately, there is (at least explicitly) no such scale-space extrema operations in existing CNNs.
Our motivation is to study the possibility of leveraging good properties of SIFT to renovate CNN  architectures towards better accuracy and robustness.

In this paper, we borrow the scale-space extrema idea from SIFT, and propose extreme value preserving networks (EVPNet) to separate robust features from non-robust ones, with three novel architecture components to model the extreme values: (1) parametric DoG (pDoG) to extract extreme values in scale-space for deep networks, (2) truncated ReLU (tReLU) to suppress noise or non-stable extrema and (3) projected normalization layer (PNL) to mimic PCA-SIFT \cite{ke2004pcasift} like feature normalization. pDoG and tReLU are combined into one block named EVPConv, which could be used to replace all $k\times k$ ($k>1$) conv-layers in existing CNNs.
We conduct comprehensive experiments and ablation studies to verify the effectiveness of each component and the proposed EVPNet. \autoref{fig:cmp} illustrates a comparison of responses for standard convolution + ReLU and EVPConv in ResNet-50 trained on ImageNet, and shows that the proposed EVPConv produces less noises and more responses around object boundary than standard convolution + ReLU,
which demonstrates the capability of EVPConv to separate robust features from non-robust ones. Our major contributions are:
\begin{itemize}[itemsep= -1pt,topsep = -1pt,partopsep=-1pt]
\item To the best of our knowledge, we are the first to explicitly separate robust features from non-robust ones in deep neural networks from an architecture design perspective.
\item We propose three novel network architecture components to model extreme values in deep networks, including parametric DoG, truncated ReLU, and projected normalization layer, and verify their effectiveness through comprehensive ablation studies.
\item We propose extreme value preserving networks (EVPNets) to combine those three novel components,
which shows to be not only more accurate, but also more robust to a set of adversarial attacks (FGSM, PGD, etc) even for clean models without adversarial training on a variety of datasets like CIFAR, SVHN, and ImageNet.
\end{itemize}

% \vspace{-1.5ex}
\section{Related Work}
% \vspace{-1.5ex}
\noindent \textbf{Robust visual features.}
Most traditional robust visual feature algorithms like SIFT \cite{lowe2004sift} and SURF \cite{bay2006surf} are based on the scale-space theory \cite{lindeberg1994scale},  while there is a close link between scale-space theory and biological vision \cite{lowe2004sift}, since many scale-space operations show a high degree of similarity with receptive field profiles recorded from the mammalian retina and the first stages in the visual cortex. For instance, DoG computes the difference of two Gaussian blurred images and is believed to mimic the neural processing in the retina \cite{young1987dog}.
SIFT is one such kind of typical robust visual features, which consists of 4 major stages: (1) scale-space extrema detection with DoG operations; (2) Keypoints localization by their stability; (3) Orientation and scale assignment based on primary local gradient direction; (4) Histogram based keypoint description.
We borrow the scale-space extrema idea from SIFT, and propose three novel and robust architecture components to mimic key stages of SIFT.

\textbf{Robust Network Architectures.}
Many research efforts have been devoted to network robustness especially on defending against adversarial attacks as summarized in \cite{akhtar2018threat}. However, there are very limited works that tackles this problem  from a network architecture design perspective.
A major category of methods~\cite{liao2018guided,xie2018denoising,menet} focus on designing new layers to perform denoising operations on the input image or the intermediate feature maps. Most of them are shown effective on black-box attacks, while are still vulnerable to white-box attacks. Non-local denoising layer proposed in \cite{xie2018denoising} is shown to improve robustness to white-box attack to an extent with adversarial training~\cite{madry2018towards}. Peer sample information is introduced in \cite{svoboda2019peernet} with a graph convolution layer to improve network robustness. Biologically inspired protection \cite{nayebi2017biologically} introduces highly non-linear saturated activation layer to replace ReLU layer, and demonstrates good robustness to  adversarial attacks, while similar higher-order principal is also used in \cite{krotov2018dense}. However, these methods still lack a systematic architecture design guidance, and many~\cite{svoboda2019peernet,nayebi2017biologically} are not robust to iterative attack methods like PGD under clean model setting. In this work, inspired by robust visual feature SIFT, we are able to design a series of innovative architecture components systematically for improving both model accuracy and robustness.

We should stress that extreme value theory is a different concept to scale-space extremes, which tries to model the extreme in data distribution, and is used to design an attack-independent metric to measure robustness of DNNs \cite{weng2018evaluating} by exploring input data distribution.

% \vspace{-1.5ex}
\section{Preliminary}
% \vspace{-1.5ex}
\noindent \textbf{Difference-of-Gaussian.} Given an input image $I_{0}$ and Gaussian kernel $G(x,y,\sigma)$ as below
\begin{equation}
% \small
G(x,y,\sigma) = \frac{1}{2\pi\sigma^{2}} e^{-(x^2+y^2)/2\sigma^2},
\end{equation}
where $\sigma$ denotes the variance. Also, difference of Gaussian (DoG) is defined as
\begin{equation}
% \small
\label{eq:dog}
D(x, y, \sigma) = G(x,y,\sigma)\otimes I_1 - G(x,y,\sigma)\otimes I_0,
\end{equation}
where $\otimes$ is the convolution operation, and $I_1 = G(x,y,\sigma)\otimes I_0$.
Scale-space DoG repeatedly convolves input images with the same Gaussian kernels, and produces difference-of-Gaussian images by subtracting adjacent image scales.
Scale-space extrema (maxima and minima) are detected in DoG images by comparing a pixel to its 26 neighbors in 3$\times$3 grids at current and two adjacent scales \cite{lowe2004sift}.

\noindent \textbf{Adversarial Attacks.} We use $h(\cdot)$ to denote the softmax output of classification networks, and $h^{c}(\cdot)$ to denote the prediction probability of class $c$. Given a classifier $h(\mathbf{x})=y$, adversarial attack aims to find $\mathbf{x}^{\text{adv}}$ such that the output of classifier deviates from the true label $y$: $\max_{i} h^{i}(\mathbf{x}^{\text{adv}})\neq y$ while closing to the original input: $||\mathbf{x} - \mathbf{x}^{\text{adv}}||\le \epsilon$. Here $||\cdot||$ refers to a norm operator, i.e. $L_{2}$ or $L_{\infty}$.

\noindent \textbf{Attack Method.}
The most simple adversarial attack method is Fast Gradient Sign Method (FGSM)~\cite{ian2015explain}, a single-step method which takes the sign of the gradient on the input as the direction of the perturbation. $L(\cdot,\cdot)$ denotes the loss function defined by cross entropy.  Specifically, the formation is as follows:
\begin{equation}
% \small
\mathbf{x}^{\text{adv}} = \mathbf{x} + \epsilon \cdot \text{sign} (\nabla_{\mathbf{x}}{L(h(\mathbf{x}), y)}),
\end{equation}
where $\mathbf{x}$ is the clean input, $y$ is the label. $\epsilon$ is the norm bound ($||\mathbf{x} - \mathbf{x}^{\text{adv}}||\le \epsilon$, i.e. $\epsilon$-ball) of the adversarial perturbation.
Projected gradient descent (PGD) iteratively applies FGSM with a small step size $\alpha$ \cite{kurakin2016adversarial,madry2018towards} with formulation as below:
\begin{equation}
% \small
\mathbf{x}^{\text{adv}}_{\text{i+1}} = \text{Proj}(\mathbf{x_{i}^{\text{adv}}} + \alpha \cdot \text{sign} (\nabla_{\mathbf{x}}{L(h(\mathbf{x}_{i}^{\text{adv}}), y)})),
\end{equation}
where $i$ is the iteration number, $\alpha = \nicefrac{\epsilon}{T}$ with $T$ being the number of iterations. `Proj' is the function to project the image back to $\epsilon$-ball every step.
Some advanced and complex attacks are further introduced in DeepFool~\cite{deepfool}, CW~\cite{CW2017}, MI-FGSM~\cite{dong2018boosting}.

\noindent \textbf{Adversarial Training} aims to inject adversarial examples into training procedure so that the trained networks can learn to classify adversarial examples correctly. Specifically, adversarial training solves the following empirical risk minimization problem:
\begin{equation}
\arg\min_{h\in H} E_{(\mathbf{x},y)~D}[\max_{\mathbf{x}^{*}\in A(\mathbf{x})} L(h(\mathbf{x}^{*}), y)],
\end{equation}
where $A(\mathbf{x})$ denotes the area around $\mathbf{x}$ bounded by $L_{\infty}$/$L_{2}$ norm $\epsilon$, and $H$ is the hypothesis space.
This paper employs both FGSM and PGD to generate adversarial examples for adversarial training.

\begin{figure*}[]
    \centering
    \subfigure[EVPConv]{\label{fig:evpconv}\includegraphics[width=0.47\linewidth]{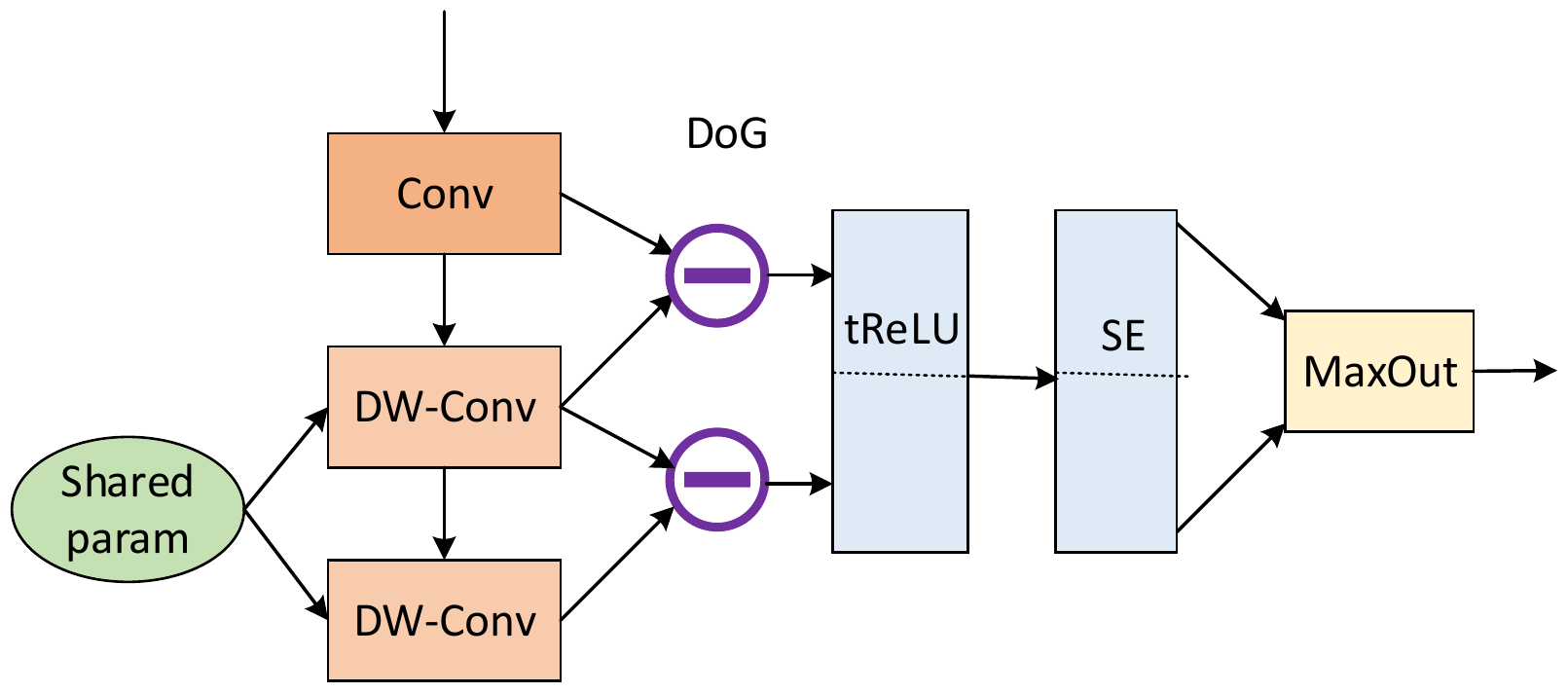}}
    \hspace{1ex}
    \subfigure[tReLU]{\label{fig:trelu}\includegraphics[width=0.28\linewidth]{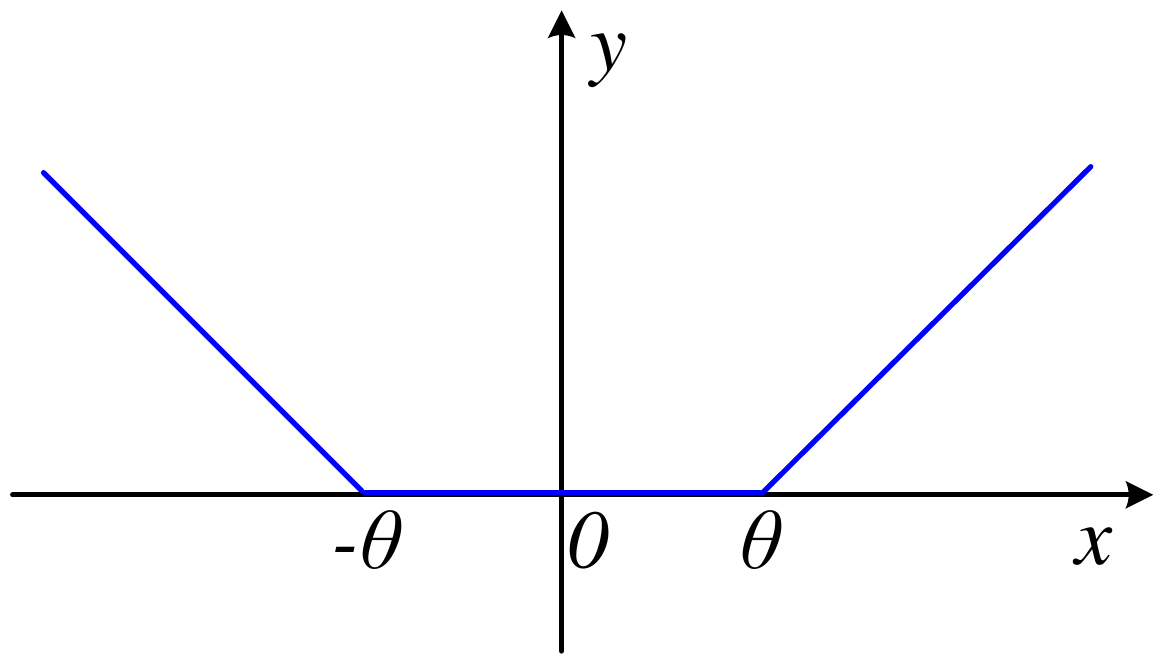}}
    \hspace{1ex}
    \subfigure[PNL]{\label{fig:pln}\includegraphics[height=0.19\linewidth]{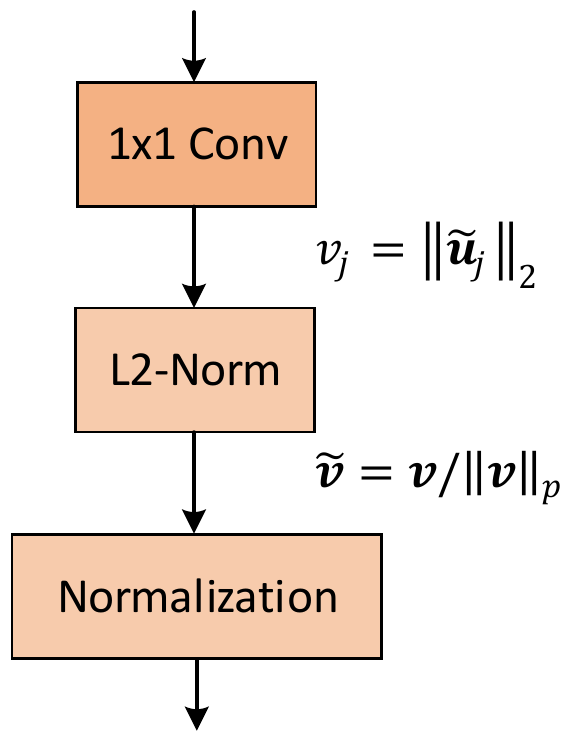}}
    \caption{Novel components in EVPNet (a) Overall architecture of EVPConv; (b) plot of truncated ReLU (tReLU) function; (c) projected normalization layer (PNL). Here $\|\cdot\|_p$ means $\ell_p$ norm.}
    \label{fig1}
\end{figure*}

\section{Method}
Inspired by traditional robust visual feature SIFT, this paper aims to improve model accuracy and robustness by introducing three novel network architecture components to mimic some key components in SIFT: parametric DoG (pDoG), truncated ReLU (tReLU), and projected normalization layer (PNL). Combining pDoG and tReLU constructs the so-called extreme value preserving convolution (EVPConv) block as shown in Figure~\ref{fig:evpconv}, which can be used to replace all $k\times k$ ($k > 1$) conv-layers in existing CNNs. PNL is a new and robust layer plugged in to replace global average pooling (GAP) layer as shown in Figure~\autoref{fig:pln}. A network with all the three components is named as extreme value preserving network (EVPNet). In the following, we will describe these three components in details separately, and elaborate on how they are used to construct the EVPConv block and EVPNet.

\subsection{Three Basic Components to mimic SIFT}\label{trelu}
\textbf{Parametric DoG (pDoG)} is a network component we design to mimic DoG operation.
Recall DoG in \autoref{eq:dog}, it repeatedly convolves input images with the \textbf{same} Gaussian kernel in which kernel size $\sigma$ is designable, and then computes the differences for adjacent Gaussian blurred images.
For CNNs, we mimic DoG with two considerations.
\textit{First}, we replace the Gaussian kernel with a learnable convolutional filter.
Specifically, we treat each channel of feature map separately as one image, and convolve it with a learnable $k\times k$ kernel to mimic Gaussian convolution.
Note that the learnable kernel is not required to be symmetric since some recent evidences show that non-symmetric DoG may perform even better \cite{einevoll2012extended,winnemoller2011xdog}. Applying the procedure to all the feature-map channels is equal to a depth-wise (DW) convolution \cite{howard2017mobilenets}.
\textit{Second}, we enforce successive depth-wise convolutions in the same block with shared weights since traditional DoG operation uses the same Gaussian kernel.
As CNNs produce full scale information at different stages with a series of convolution and downsampling layers, each pDoG block just focuses on producing extrema for current scale, while not requiring to produce full octave extrema like SIFT.
The shared DW convolution introduces minimum parameter overhead, and avoid ``extrema drift'' in the pDoG space so that it may help finding accurate local extrema.
Formally, given input feature map $\mathbf{f}_0$, a minimum of two successive depth-wise convolution is applied as
\begin{equation}
    \mathbf{f}_1 = DW(\mathbf{f}_0; \mathbf{w}); \quad \mathbf{f}_2 = DW(\mathbf{f}_1; \mathbf{w}),
\end{equation}
where $DW(;)$ is depth-wise convolution with $\mathbf{w}$ as the shared weights. pDoG is thus computed as
\begin{equation}
    \mathbf{d}_0 = \mathbf{f}_1 - \mathbf{f}_0; \quad \mathbf{d}_1 = \mathbf{f}_2 - \mathbf{f}_1 = DW(\mathbf{f}_1; \mathbf{w}) - DW(\mathbf{f}_0; \mathbf{w}).
\end{equation}
It is worth noting that the if the minus sign is merged into weight, we will get different filters ($\bf w_1 =w, w_2 = -w$).
\textbf{Minus operator} is thus introduced into deep neural networks.

Following SIFT, we compute local extrema Qacross the pDoG images using maxout operations \cite{goodfellow2013maxout}:
\begin{equation} \label{eq:z}
\footnotesize
    \mathbf{z}_0 = \max(\mathbf{d}_0, \mathbf{d}_1); \quad \mathbf{z}_1 = -\min(\mathbf{d}_0, \mathbf{d}_1) = \max(-\mathbf{d}_0, -\mathbf{d}_1).
\end{equation}
Note we do not compute local extrema in $3\times 3$ spatial grids as in SIFT since we do not require keypoint localization in CNNs.
Finally, to keep the module compatible to existing networks, we need ensure the output feature map to be of the same size (number of channels and resolution).
Therefore, a maxout operation is taken over to merge two feature maps and obtain the final output of this block:
\begin{equation}\label{eq:o}
\mathbf{o} = \max(\mathbf{z}_0, \mathbf{z}_1).
\end{equation}
\textbf{Truncated ReLU (tReLU)}.
The pDoG block keeps all the local extrema in the DoG space, while many local extrema are unstable because of small noise and even contrast changes. SIFT adopts a local structure fitting procedure to reject those unstable local extrema. To realize similar effect, we propose truncated ReLU (tReLU) to suppress non-robust local extrema.
The basic idea is to truncate small extrema which correspond to noise and non-stable extrema in the pDoG space. This can be implemented by modifying the commonly used ReLU function as
\begin{equation}
  y = \left\{
    \begin{array}{rl}
      |x| & \text{if }  |x| \ge \theta,\\
      0 & \text{if } |x| < \theta,
    \end{array} \right.
\end{equation}
where $\theta$ is a learnable parameter. Note that this function is discontinued at $x$ = $\pm \theta$. We make a small modification to obtain a continuous version for easy training as below
\begin{equation}
  y = tReLU(x) = \left\{
    \begin{array}{rl}
      |x|-\theta & \text{if }  |x| \ge \theta,\\
      0 & \text{if } |x| < \theta.
    \end{array} \right.
\end{equation}
Figure~\autoref{fig:trelu} plots the tReLU function. Different from the original ReLU, tReLU introduces a threshold parameter $\theta$ and keeps elements with higher magnitude.
$\theta$ is designed as a block-level parameter (i.e., each block has one threshold) by default.

tReLU is combined with pDoG not only to suppress non-robust extrema, but also to simplify the operations.
When combining \autoref{eq:z} and \autoref{eq:o} together, there is nested maxout operation which satisfies the commutative law, so that we could rewrite $\mathbf{z}_0$ and $\mathbf{z}_1$ as
\begin{equation} \label{eq:z1}
    \mathbf{z}_0 = \max(\mathbf{d}_0, -\mathbf{d}_0) = |\mathbf{d}_0|; \quad \mathbf{z}_1 = \max(\mathbf{d}_1, -\mathbf{d}_1)=|\mathbf{d}_1 |,
\end{equation}
where $|\cdot|$ is element-wise absolute operation. With tReLU to suppress non-robust features, we have
\begin{equation} \label{eq:ztrelu}
    \mathbf{z}_0 = tReLU(\mathbf{d}_0); \quad \mathbf{z}_1 = tReLU(\mathbf{d}_1).
\end{equation}
Hence, in practice, we use \autoref{eq:ztrelu} instead of \autoref{eq:z} to compute $\mathbf{z}_0$ and $\mathbf{z}_1$. Note that tReLU does improve robustness and accuracy for pDoG feature maps, while providing no benefits when replacing ReLU in standard CNNs according to our experiments (see~\autoref{ablate1}).

\textbf{Projected Normalization Layer (PNL)}.
SIFT~\cite{lowe2004sift} computes gradient orientation histogram followed by $L2$ normalization to obtain final feature representation.
This process does not take gradient pixel relationship into account. PCA-SIFT \cite{ke2004pcasift} handles this issue by projecting each local gradient patch into a pre-computed eigen-space using PCA.
We borrow the idea from PCA-SIFT to build projected normalization layer (PNL) to replace global average pooling (GAP) based feature abstraction in existing CNNs.
Suppose the feature-map data matrix before GAP is $\mathcal{X} \in \mathbb{R}^{d\times c}$, where $d = w\times h$ is the feature map resolution, and $c$ is the number of channels, we obtain column vectors $\{\mathbf{x}_i \in \mathbb{R}^{c}\}_{i=1}^d$ from $\mathcal{X}$ to represent the $i$-th pixel values from all channels. The PNL contains three steps:
\begin{itemize}[itemsep= -1pt,topsep = -1pt,partopsep=-1pt]
\item[(1)] We add a $1\times 1$ conv-layer (not required for network with $1\times 1$ conv-layer before GAP), which can be viewed as a PCA with learnable projection matrix $\mathcal{W}\in \mathbb{R}^{c\times p}$. The output is $\mathbf{u}_i = \mathcal{W}^T \mathbf{x}_i$, where $\mathbf{u}_i \in \mathbb{R}^p$ further forms a data matrix $\mathcal{U}\in \mathbb{R}^{d\times p}$.
\item[(2)] We compute $L2$ norm for each row vectors $\{\tilde{\mathbf{u}}_j\in \mathbb{R}^d\}_{j=1}^p$ of $\mathcal{U}$, to obtain a vector $\mathbf{v} = (v_1, \cdots, v_p)$ with $v_j = \|\tilde{\mathbf{u}}_j\|_2$.
\item[(3)] To eliminate contrast or scale impact, we normalize $\mathbf{v}$ to obtain $\tilde{\mathbf{v}} = \mathbf{v}/\|\mathbf{v}\|_p$, while
$\|\cdot\|_p$ means a $\ell_p$ norm.
Vector $\tilde{\mathbf{v}}$ is fed into classification layer for prediction purpose.
\end{itemize}
It is interesting to note that PNL actually computes a second order pooling similar as \cite{gao2019global,yu2018smsop}.
Suppose $\mathbf{w}_j \in$ $\mathbb{R}^c$ is the $j$-th row of $\mathcal{W}$, $v_j$ in step-2 can be rewritten as
\begin{equation}
\begin{split}
   v_j &= \sqrt{\sum\nolimits_{i=1}^d  (\mathbf{w}_j^T \mathbf{x}_i)^2} \\&= \sqrt{\sum\nolimits_{i=1}^d  \mathbf{w}_j^T \mathbf{x}_i \mathbf{x}^T_i \mathbf{w}_j}
   \\&= \sqrt{\mathbf{w}_j^T  (\sum\nolimits_{i=1}^d  \mathbf{x}_i \mathbf{x}^T_i) \mathbf{w}_j} \\&= \sqrt{\mathbf{w}_j^T  A \mathbf{w}_j}
\end{split}
\end{equation}
where $A = \sum\nolimits_{i=1}^d  \mathbf{x}_i \mathbf{x}^T_i$ is an auto-correlation matrix. Figure~\autoref{fig:pln} illustrates the PNL layer.
Theoretically, GAP produces a hyper-cube, while PNL produces a hyper-sphere.
This is beneficial for robustness since hyper-sphere is more smooth, and a smoothed surface is proven more robust \cite{cohen2019certified}.
Our experiments also verify this point (see~\autoref{ablate1}).

\subsection{EVPConv and EVPNet}
With these three novel components, we can derive a novel convolution block named EVPConv, and the corresponding networks EVPNet. In details, EVPConv starts from the pDoG component, and replaces \autoref{eq:z} with tReLU as in \autoref{eq:ztrelu}. In SIFT, the contribution of each pixel is weighted by the gradient magnitude. This idea can be extended to calibrate contributions from each feature-map channel.
Fortunately, Squeeze-and-Excitation (SE) module proposed in \cite{hu2018senet} provides the desired capability. We thus insert the SE block after tReLU, and compute the output of EVPConv as:
\begin{equation}
\begin{split}
    &\mathbf{o} = \max(\mathbf{s}_0, \mathbf{s}_1),\\&~~where~~ (\mathbf{s}_0, \mathbf{s}_1) = SE(concat(\mathbf{z}_0, \mathbf{z}_1))
\end{split}
\end{equation}
where \textit{concat}$(\cdot)$ means concatenating $\mathbf{z}_0$ and $\mathbf{z}_1$ together for a unified and unbiased calibration, $SE(\cdot)$ is the SE module, $\mathbf{s}_0$ and $\mathbf{s}_1$ are the calibration results corresponding to $\mathbf{z}_0$ and $\mathbf{z}_1$, and $\max$ denotes an element-wise maximum operation. Figure~\autoref{fig:evpconv} illustrates the overall structure of EVPConv.

EVPConv can be plugged to replace any $k\times k$ ($k>$1) conv-layers in existing CNNs, while the PNL layer can be plugged to replace the GAP layer for feature abstraction. The network consisting of both EVPConv block and the PNL layer is named as EVPNet.
The EVPConv block introduces very few additional parameters: $\mathbf{w}$ for shared depth-wise convolution, $\theta$ for tReLU and parameters for SE module, which amounts to  7$\sim$20\% extra parameters (see \autoref{app1}). It also increases theoretic computing cost 3$\sim$10\% for a bunch of parameter-free operations like DoG and maxout. However, the added computing cost is non-negligible in practice (2$\times$ slower according to our training and inference experiments) due to more memory cost for additional copy of feature-maps.
Near memory computing architecture \cite{singh2019near} may provide efficient support for this new computing paradigm.

\section{Experiments}
\subsection{Experiments on CIFAR-10}
\textbf{Experimental Setup.}
We first evaluate the proposed network components and EVPNet on CIFAR-10, which is a widely used image classification dataset containing $60,000$ images of size 32$\times$32 with $50,000$ for training and 10,000 for testing.
Supplementary provides extra experiments on SVHN.
\begin{table*}[t]
\centering
\caption{Ablation study results on CIFAR-10, attacking with $\epsilon$=8. `PGD-$N$-$s$' denotes PGD attack for $N$ iterations of step $s$ pixels.
}
\vspace{1ex}
\label{ablate1}
\begin{tabular}{c|ccc|cccccc}
\hline
Network &  +pDoG & +tReLU   & +PNL  & Clean &  FGSM & PGD-10-1 & PGD-10-2 & DeepFool & CW \\ \hline
  \multirow{8}{*}{SE-ResNet-20}     &      &    &   &  91.68       &  24.23   & 3.03 & 0.00 &20.19 &0.57  \\
       &     \checkmark   &    &            & 92.55    & 31.14 &     3.37 & 0.00   & 29.33 & 0.60 \\
        &       &  \checkmark &  &   91.38  & 25.37 & 2.79 & 0.00  & 23.41 & 0.23\\
       &        &  & \checkmark         & 90.92  & 40.66      & 13.16 & 1.66  & 31.86 & 1.02\\
       &    \checkmark   &    \checkmark  &      & 91.98    & 37.46 & 4.64    & 0.01  & 36.54& 0.98\\
       &      \checkmark &   & \checkmark          & 92.62  &   60.12    &  28.03  & 6.37 & 39.28 & 4.56\\
       &      \checkmark   & \checkmark   &\checkmark  &   \textbf{92.85}   &   \textbf{66.21}    &  \textbf{41.06} & \textbf{12.19}  &\textbf{40.28} &\textbf{6.78}  \\    \hline
\end{tabular}
\end{table*}

We introduce our novel components into the well-known and widely used ResNet, and compare to the basic model on both clean accuracy and adversarial robustness. As the EVPConv block contains a SE module, to make a fair comparison, we set SE-ResNet as our comparison baseline. In details, we replace the input conv-layer and the first $3\times 3$ conv-layer in the residual block with EVPConv, and replace the GAP layer with the proposed PNL layer. All the networks are trained with SGD using momentum 0.9, 160 epochs in total. The initial learning rate is 0.1, divided by 10 at 80 and 120 epochs.
Results can be compared to \cite{he2016deep,densenet} due to different training epochs. For tReLU, the channel-level parameter $\theta$ is initialized by uniformly sampling from $[0,1]$.

In this work, we consider adversarial perturbations constrained under $l_{\infty}$ norm. The allowed perturbation norm $\epsilon$ is 8 pixels~\cite{madry2018towards}. We evaluate non-targeted attack adversarial robustness in three settings: normal training, FGSM adversarial training~\cite{ian2015explain,tramer2017ensemble} and PGD adversarial training~\cite{madry2018towards}.
During adversarial training, we use the predicted label to generate adversarial examples to prevent label leaking  effect~\cite{kurakin2017adversarial}. To avoid gradient masking~\cite{tramer2017ensemble}, we use R-FGSM for FGSM adversarial training, which basically starts from a random point in the $\epsilon$ ball. Following \cite{madry2018towards}, during training, PGD attacks generate adversarial examples by 7 PGD iterations with 2-pixel step size starting from random points in the allowed $\epsilon$ ball. We report accuracy on both whitebox and blackbox attack.
We evaluate a set of well-known whitebox attacks, including FGSM, PGD, DeepFool, CW. We use `PGD-$N$' to denote attack with $N$ PGD iterations of step size 2 pixels by default.
Specifically, we compare results for PGD-10 and PGD-40. For blackbox attack, we choose VGG-16 as the source model which is found by \cite{su2018is} to exhibit high adversarial transferability, and choose FGSM as the method to generate adversarial examples from VGG-16 as it is shown to lead to better transferability~\cite{su2018is}.

\noindent \textbf{Ablation Study.} This part conducts a thorough ablation study to show the effectiveness of each novel architecture component and how they interact to provide strong adversarial robustness.
We conduct experiments on CIFAR-10 with SE-ResNet-20, which contains one input conv-layer, three residual stage each with two bottleneck residual blocks, and GAP layer followed by a classification layer. We evaluate the accuracy and robustness for all the possible combinations of the proposed three components under the normal training setting. For PGD attack, we use two step sizes: 1 pixel as in \cite{xie2018denoising} and 2 pixels as in \cite{madry2018towards}.
\autoref{ablate1} lists full evaluation results of each tested model. Several observations can be made from the table:(1) Solely adding pDoG or PNL layer leads to significant robustness improvement. pDoG even yields clean accuracy improvement, while PNL yields slightly clean accuracy drops; (2) tReLU does not bring benefit for standard convolution, while yields notable improvement on both clean accuracy and adversarial accuracy, when combining with pDoG. That verifies our previous claim that tReLU is suitable to work for the DoG space; (3) Combining all the 3 components together obtains the best adversarial robustness, while still reach 1.2\% clean accuracy improvement over the baseline without these three components.
Based on these observations, we incorporate all the three components into the CNNs to obtain the so-called EVPNet for the following experiments if not explicitly specified.
As 2-pixels PGD attack is much stronger than 1-pixel PGD attack, we use it as default in the following studies.
\begin{table*}[t]
\centering
\caption{Comparison results on CIFAR-10 at different training settings and different networks, attack with $\epsilon=8$.}
\vspace{1ex}
\label{base}
\begin{tabular}{c|c|cccccccc}
\hline
Network & Training & \multicolumn{1}{c}{Model} & Clean  & FGSM  & PGD-10  & PGD-40 & DeepFool & CW & Blackbox \\ \hline
\multirow{6}{*}{SE-ResNet-20} & \multirow{2}{*}{Normal} & \multicolumn{1}{c}{Baseline} & 91.68   & 24.23 & 0.00   & 0.00  & 20.19 & 0.57   &78.32    \\
                           &  & EVPNet     & \textbf{92.85}  &\textbf{66.21}   &\textbf{12.19}  & \textbf{6.84}  & \textbf{40.28} & \textbf{6.78} & \textbf{79.31}  \\ \cline{2-10}
                           & \multirow{2}{*}{FGSM}  & \multicolumn{1}{c}{Baseline} &  84.84 &   62.17  & 29.41 & 27.74  &60.88 & 25.54   & 81.81  \\
                           &  & EVPNet  & \textbf{84.94}  &  \textbf{64.90} & \textbf{34.10} & \textbf{31.75} & \textbf{62.74} &\textbf{29.87} &  \textbf{82.50} \\ \cline{2-10}
                           & \multirow{2}{*}{PGD}  & \multicolumn{1}{c}{Baseline} &  84.02 & 64.39  & 36.23  &  34.84  & 63.01 & 32.59  &  82.71  \\
                           &   & EVPNet  &   \textbf{84.30}  &  \textbf{65.90} &  \textbf{38.60}  & \textbf{36.94} & \textbf{64.23} & \textbf{34.14}  &  \textbf{83.83}  \\ \hline
\multirow{6}{*}{SE-ResNet-56} & \multirow{2}{*}{Normal} & \multicolumn{1}{c}{Baseline} &  \textbf{94.20}   &  37.18   &    0.00 &  0.00  &24.33 & 0.97 &83.42   \\
                           &  & EVPNet &   93.80  & \textbf{74.05} & \textbf{29.39} & \textbf{7.43} & \textbf{44.59} & \textbf{8.33} &  \textbf{84.04}  \\ \cline{2-10}
                           & \multirow{2}{*}{FGSM}  & \multicolumn{1}{c}{Baseline} & 87.50 & 67.88 & 35.84  & 33.87   & 62.34 & 34.54 &   86.17 \\
                           &  & EVPNet  &\textbf{89.61} & \textbf{69.93} &    \textbf{41.79} &  \textbf{35.63} & \textbf{63.45} &\textbf{35.42} & \textbf{87.57}  \\ \cline{2-10}
                           & \multirow{2}{*}{PGD} & Baseline   & 86.86  & 69.45 &  39.90 & 39.90 & 64.22 & 38.90  &  87.33 \\
                            & & EVPNet & \textbf{90.13} & \textbf{71.33} & \textbf{40.81} & \textbf{40.80} & \textbf{66.11} & \textbf{39.10} &\textbf{89.00} \\\hline
\end{tabular}
\end{table*}

\begin{figure*}[]
 \centering
\begin{minipage}{0.49\linewidth}
 \centering
 \subfigure[]{\label{arch-search-a1}\includegraphics[width=0.48\linewidth]{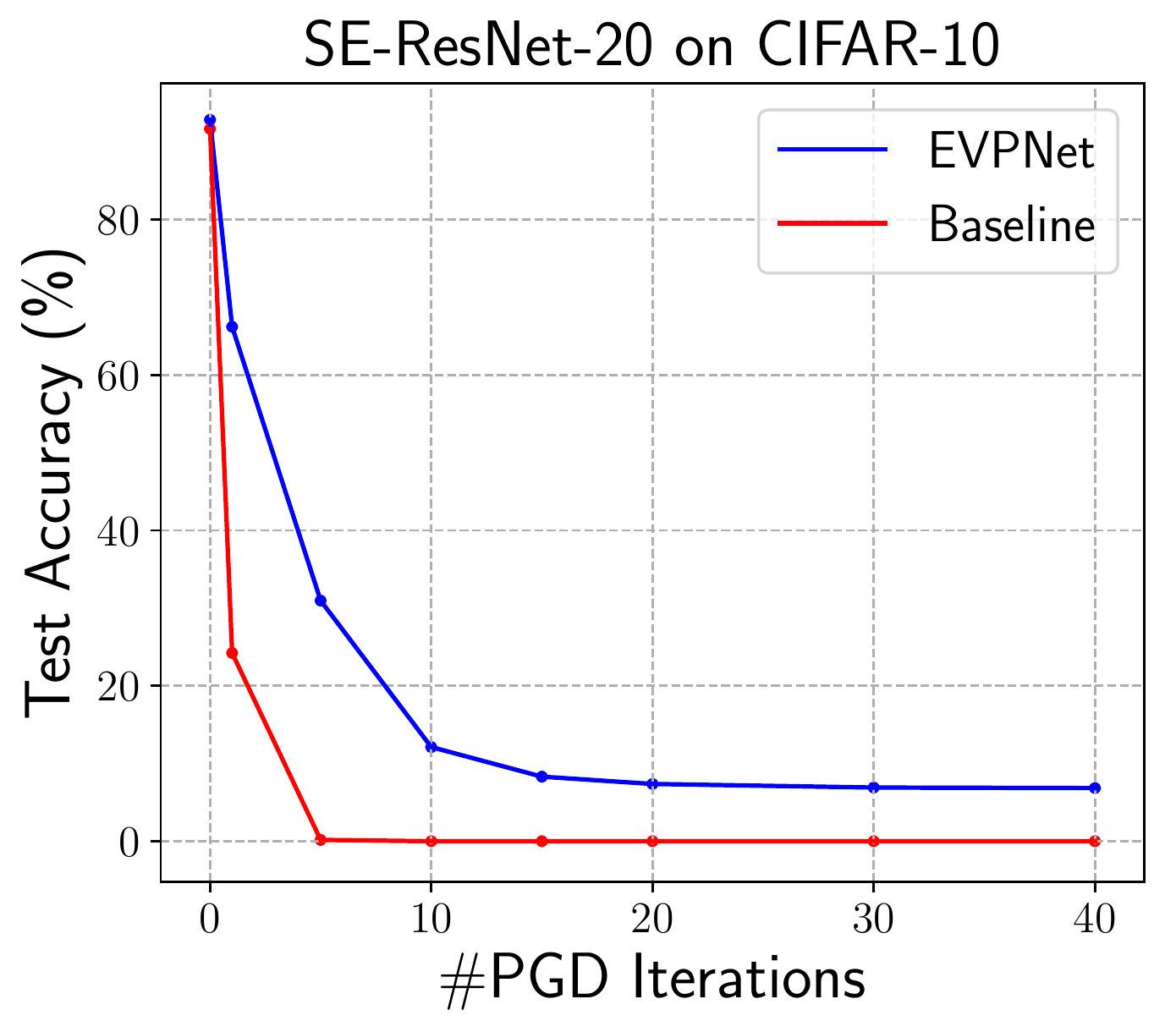}}
  \subfigure[]{\label{arch-search-a2}\includegraphics[width=0.48\linewidth]{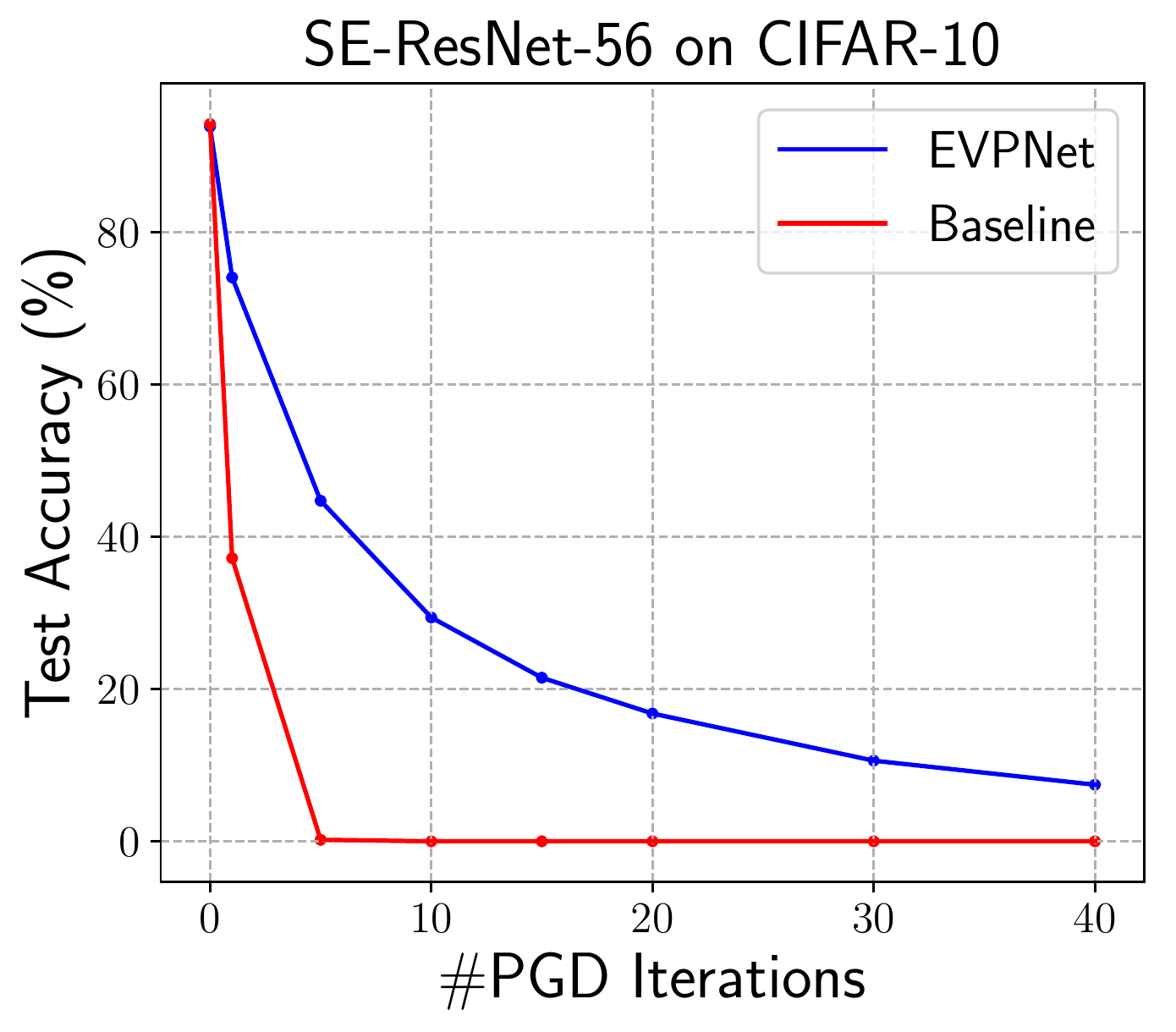}}
\caption{Test accuracy \textit{vs} the number of PGD iterations for both networks on two datasets. Better visualized in colors.}
\label{fig2}
\end{minipage}
\hspace{1ex}
\begin{minipage}{0.49\linewidth}
 \centering
 \subfigure[]{\label{arch-search-a3}\includegraphics[width=0.48\linewidth]{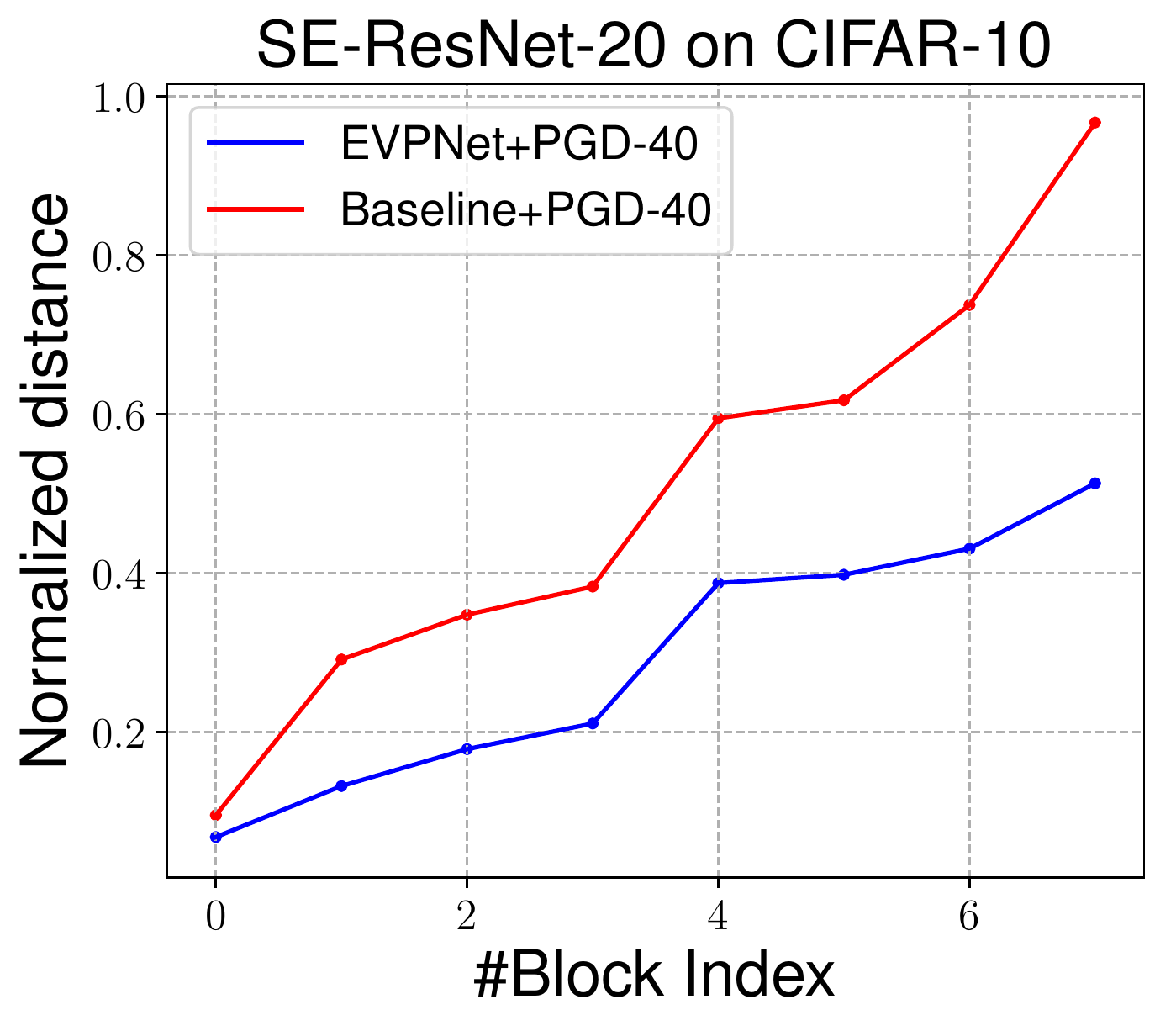}}
  \subfigure[]{\label{arch-search-a4}\includegraphics[width=0.48\linewidth]{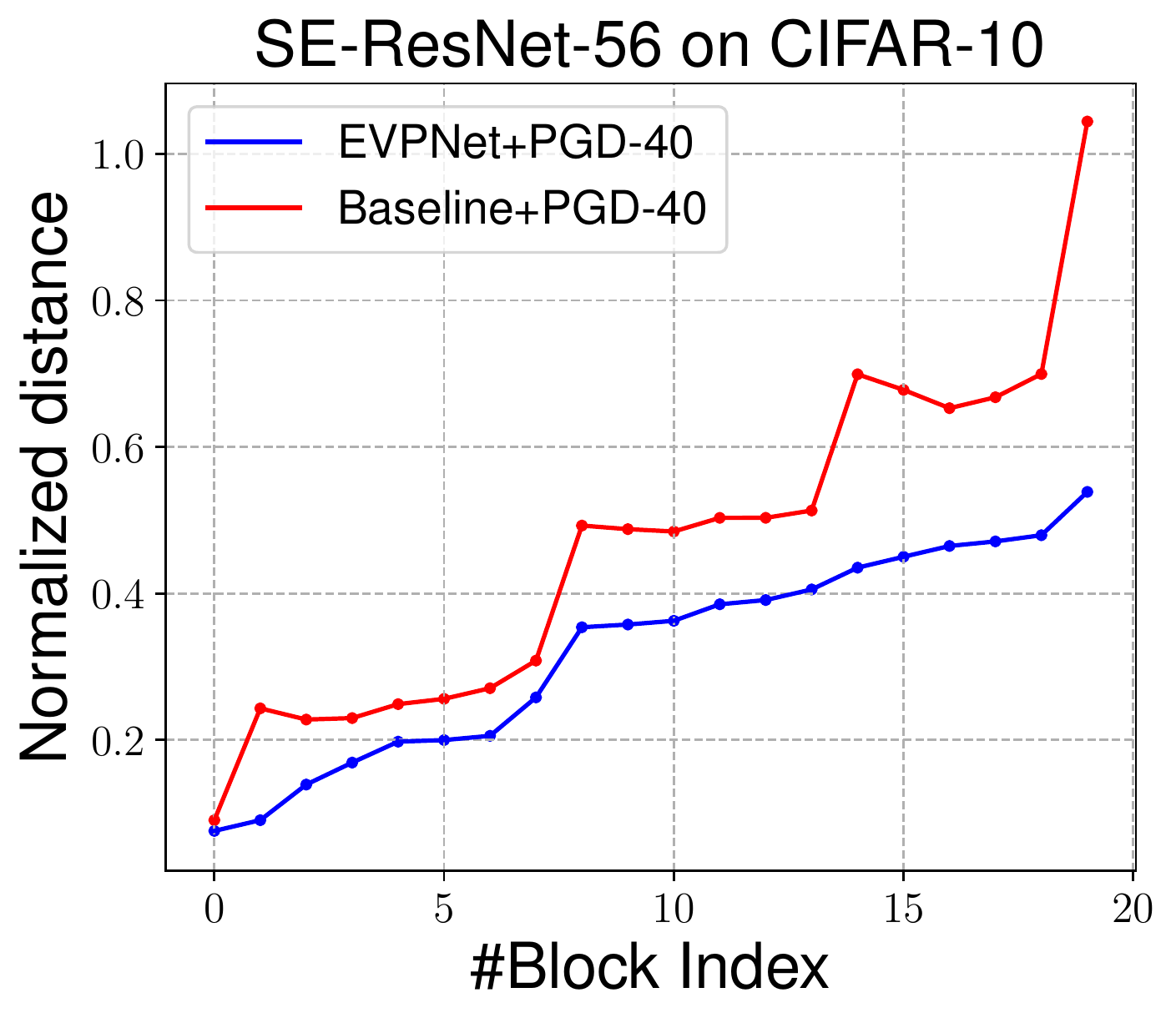}}
\caption{
Normalized adversarial-benign distance for feature responses between adversarial and benign examples at different res-block.}
\label{fig3}
\end{minipage}
\end{figure*}

\textbf{Benchmark Results.}
We conduct extensive experiments on CIFAR-10 to compare the proposed EVPNet with the source networks. The two sources networks are ResNet-20 and ResNet-56. For fair comparison, we use the SE extended ResNet as our baseline. \autoref{base} lists comprehensive comparison results. We list 7 different kinds of accuracies: clean model accuracy, whilebox attack accuracies by FGSM/PGD-10/PGD-40/DeepFool/CW, and blackbox attack accuracy with adversarial examples generated by FGSM on the VGG-16 model. We can see that under normal training case, EVPNet outperforms baseline by a large margin in terms of robustness with FGSM, PGD, DeepFool, and CW attacks. Even under the strongest PGD-40 white box attack, our EVPNet still has non-zero accuracy without any adversarial training. For those cases with adversarial training, our EVPNet consistently beats baseline networks with noticeable margin.

\textbf{Analysis.}
We make further analysis to compare EVPNet to baseline networks. \textit{First}, we plot the test error at different PGD iterations for different evaluated networks under normal training case as shown in \autoref{fig2}. It can be seen that EVPNet consistently outperforms the corresponding baseline networks significantly under all PGD iterations. Some may concern that the accuracy of EVPNet on the strongest PGD-40 attack is not satisfied ($\sim$ 10\%). We argue that from three aspects:
(1) The adversarial result is remarkable as it is by the clean model without using any other tricks like adversarial training, etc. (2) The proposed components also brings consistent clean accuracy improvement even on large-scale dataset (see \autoref{app1}), while most existing methods on adversarial robustness may even yield clean accuracy drops.
(3) The methodology we developed may shed some light on future studies in network robustness and network architecture design/search.

\textit{Second}, we further investigate the error amplification effect as \cite{liao2018guided}.
Specifically, we feed both benign examples and adversarial examples into the evaluated networks, and compute the normalized $L_{2}$ distance for each res-block outputs as $\gamma = \|\mathbf{x} - \mathbf{x}'\|_2 /\|\mathbf{x}\|_2$, where $\mathbf{x}$ is the response vector of benign example, and $\mathbf{x}'$ is the response vector for the adversarial example. We randomly sample 64 images from the test set to generate adversarial examples using PGD-40. The models evaluated are trained without adversarial training.
\autoref{fig3} illustrates the results. As we can see, EVPNet has much lower average normalized distance than the baseline models almost on all the blocks.
It is interesting to see that the baseline models have a big jump for the normalized distance at the end of the networks.
This urges the adversarial learning researchers to make further investigation on the robustness of deep layers. Nevertheless, EVPNet significantly reduces the error amplification effect in this analysis.

\textit{Third}, we compare the differences on feature responses between regular convolution + ReLU and EVPConv. This comparison is made on large-scale and relative high resolution ($224\times 224$) ImageNet dataset for better illustration. We train ResNet-50 and EVPNet-50 on ImageNet, and visualize their prediction responses for the first corresponding convolution block in \autoref{fig:cmp}.
It clearly shows that EVPNet-50 gives more responses on object boundary. This demonstrates the capability of EVPConv to separate robust features from non-robust ones.

\begin{table*}[t]
\centering
\caption{Top-1 accuracy by Resnet26A/B, Resnet50, and MobileNet on ImageNet under normal training. The last two columns list the model parameter size and model computing FLOPs. }
\vspace{1.5ex}
\label{tab:imgnet}
\begin{tabular}{c|c|ccc|cc}
\hline
Network  & Model & Clean   & FGSM  & PGD-10   &\#Params(M) &FLOPs(G) \\ \hline
\multirow{3}{*}{ResNet26A} & Baseline & 57.63 & 21.45  & 0.10   &1.33  &0.261  \\
    &  SE-ResNet  & 60.11 &26.98& 0.45  &1.35 &0.262 \\
    &  EVP-ResNet  &  \textbf{63.72} & \textbf{30.23} & \textbf{6.78}  &1.65 & 0.290 \\ \hline
\multirow{3}{*}{ResNet26B} & Baseline & 72.39 &28.31 & 0.10 &15.24 &2.34 \\
    &  SE-ResNet  & 74.14 & 32.19 & 1.02  &15.58 &2.36  \\
    &  EVP-ResNet  &  \textbf{75.38} &  \textbf{38.12} & \textbf{9.45}  &17.01 & 2.41 \\ \hline
\multirow{3}{*}{ResNet50} & Baseline & 75.04 &  29.13 &  0.23  &24.36 &4.10 \\
    &  SE-ResNet  & 76.18 & 33.45  & 1.20  &24.97 &4.11  \\
    &  EVP-ResNet  &  \textbf{77.01} &  \textbf{40.12} & \textbf{8.30}  &26.75 & 4.18\\ \hline
\multirow{3}{*}{MobileNet-v1} & Baseline & 70.07 &  18.45 & 0.09   &4.04 &0.557 \\
    &  SE-MobileNet  & 71.51 & 30.92  & 0.87  &4.70 &0.560  \\
    &  EVP-MobileNet  &\textbf{72.63}   & \textbf{37.54} & \textbf{6.99}  &5.48 &0.623 \\ \hline
\end{tabular}
\end{table*}

\subsection{Experiments on ImageNet}\label{app1}
We further make experiments on large-scale ImageNet dataset.
We propose to replace the widely used ResNet architecture with the proposes components, and obtain the derived network named EVP-ResNet.
We did experiments on 3 ResNet variants: ResNet26A/26B and ResNet50.
ResNet26A/26B are two variants of ResNet18, with the same number of residual blocks (ResBlock), but replacing standard ResBlock with bottleneck ResBlock.
ResNet26A has the same output filter width in each stage as ResNet18 (i.e, [64, 64, 128, 256, 512]). Due to the bottleneck structure, ResNet26A has just about 1/8 parameters of ResNet18.
ResNet26B has the same output filter width in each stage as ResNet50 (i.e., [64, 256, 512, 1024, 2048]), which yields just slightly higher parameter size than ResNet18, but much lower than ResNet50.
Note there are two small modifications. \textit{First}, we move the SE module in SE-ResNet after the last conv-layer in the bottleneck ResBlock to after the 2nd conv-layer, which is consistent with our EVPConv block for fair comparison. \textit{Second}, for ResNet, we implement the stride = 2 convolution in downsampling ResBlock as ResNet-B structure in \cite{he2019bag}.

Besides ResNet, we also try to extend our framework on MobileNet-v1 \cite{howard2017mobilenets}, which is a plane network structure without residual connections similar to VGGNet.
We take standard MobileNet-V1 (1.0$\times$ in channel width) as baseline, and provide experiments comparison to the SENet extension and the EVPNet extension.
For our EVPNet extension, we add EVPConv module after each $3\times 3$ depth-wise convolution, and replace the GAP layer with our PNL layer.
For the SENet extension, we add SE module after each $3\times 3$ DW convolution for fair comparison.

We trained all the models (ResNet and extensions, MobileNet and extensions) by ourselves on the same GPU server with 8 GTX 2080TI GPUs with the same training configurations:
(1) the same data augmentation (the baseline augmentation used by \cite{he2019bag});(2) Nesterov Accelerated Gradient (NAG) descent as the optimizer;
(3) training 90 epochs with batchsize 256; (4) the learning rate is initialized to 0.1 and divided by 10 at the 30th, 60th, and 80th epochs.
We evaluated model performance on the validation set with just single center-crop test, without using any data augmentation in the test procedure.
Besides clean model accuracy, we also evaluate the adversarial attacks by FGSM and PGD-10 with maximum perturbation for each pixel $\epsilon=8$ under $l_{\infty}$ norm.

\autoref{tab:imgnet} lists the full comparison results as well as model parameter size and computing cost (in terms of FLOPs). It shows that EVP-ResNet gives consistent accuracy improvement over ResNet and SE-ResNet in all the three tested network architectures. EVP-ResNet26A improves ResNet26A and SE-ResNet26A by 6.09\% and 3.61\% respectively in absolute accuracy. EVP-ResNet50 improves ResNet50 and SE-ResNet50 by 1.97\% and 0.83\% respectively in absolute accuracy.

In terms of parameter size, EVP-ResNet brings 7$\sim$20\% additional parameters to the ResNet counterparts (larger models with relative less increasing). In terms of computing FLOPs, EVP-ResNet increases 3$\sim$10\% for a bunch of parameter-free operations like DoG and maxout. However, the added computing cost is non-negligible in practice (2$\times$ slower in our training and inference experiments) due to more memory cost for additional copy of feature-maps.

In the adversarial robustness evaluation, EVP-ResNet still consistently beats the ResNet and SE-ResNetcounterparts.
For the strongest PGD-10 attacks, all ResNet and SE-ResNet models drop top-1 accuracy to near zero, while the EVP-ResNet variants keep 6$\sim$10\% top-1 accuracy. The gap in FGSM attacks is even larger. This improvement is remarkable considering that it is by clean model without adversarial training. Better results could be achieved when combining with adversarial training.

For the MobileNet case, when comparing EVP-MobileNet to the baseline counterparts, we also observed significant clean accuracy improvement and adversarial robustness improvement as shown in \autoref{tab:imgnet}.
In summary, this experiment demonstrates that the proposed EVPNet has good generalization capability on large-scale datasets and networks.

\section{Conclusion}
This paper mimics good properties of robust visual feature SIFT to renovate CNN architectures with some novel architecture components, and proposes
the extreme value preserving networks (EVPNets). Experiments demonstrate that EVPNets can achieve similar or better accuracy over conventional CNNs, while achieving significant better robustness to a set of adversarial attacks (FGSM, PGD, etc) even for clean model without using any other tricks like adversarial training.

{\small
\bibliographystyle{unsrt}
\bibliography{evpnet}
}

\newpage
\begin{appendices}

\section{Correspondences between SIFT and EVPNet modules}
Here we provide a module level correspondences between EVPNet and SIFT in \autoref{correspond}. 
We can see that most modules in SIFT have correspondences in EVPNet execept the keypoint orientation and description modlues, which are for local keypoints. 
CNNs are for image level image representation rather than for local representation. If we want to use SIFT as image level representation, we could use bag of SIFT based visual words, i.e., CNN correspond to bag of SIFT words. 
Furthermore, as an image-level representation CNNs (EVPNets) do not require keypotint orientation and scale assignment as that in SIFT.  

\begin{table*}[h]
	\centering
	\caption{Module level correspondence between SIFT and EVPNet}
	\vspace{1ex}
	\label{correspond}
	\begin{tabular}{c|c|c}
		\hline
	Module	& SIFT & EVPNet\\
		\hline
	Filters & Gaussian kernel & depthwise learnable kernel \\
		\hline
	Scale-space & DoG & pDoG \\
		\hline
	Extrema & $3\times 3\times 3$ extrema & Maxout \\
		\hline
	Calibration & gradient magnitude & SE module \\
		\hline 
	Noise stability & local fitting based outlier rejection & tReLU \\
		\hline 
	keypoint orientation & Yes & No \\
		\hline 
	Scale assignment & Yes & No \\
		\hline
	Description & Histogram of gradient or PCA-SIFT & PNL \\
		\hline
	Image-level representation & bag of SIFT based visual words & CNN \\
	    \hline
	\end{tabular}
	\vspace{-2ex}
\end{table*}

\section{Resuts on SVHN}
\autoref{base2} further lists comparison results on SVHN. Similarly, our EVPNet consistently outperforms baseline models on all the three training settings.
It is interesting to note that PGD adversarial training performs worse on PGD-10/PGD-40 attacks than FGSM adversarial training, and even much worse than normal training on this dataset. This may be due to the fact that SVHN is a shape/edge dominating digit recognition dataset, which may generate a lot of difficult adversarial samples with broken edges.
And it also coincides with the finding by \cite{baker2018deep}.
Our EVPNet shows better robustness on this dataset without adversarial training than CIFAR-10, which may suggest that EVPNet is more robust on shape/edge dominating object instances.
All these evidences prove that the proposed EVPNet is a robust network architecture.

\begin{table*}[!htbp]
	\centering
	\caption{Comparison results on SVHN at different training settings and different networks, attack with $\epsilon=8$.}
	\vspace{1ex}
	\label{base2}
	\begin{tabular}{c|c|cccccccc}
		\hline
		Network   & Training   & \multicolumn{1}{c}{Model} & Clean  & FGSM & PGD-10  & PGD-40  &DeepFool &CW & Blackbox   \\ \hline
		\multirow{6}{*}{SE-ResNet-20} & \multirow{2}{*}{Normal} & \multicolumn{1}{c}{Baseline} &  95.76   &   70.44  &   50.18 & 0.13  &  32.19 & 0.22 & 90.27 \\
		&  & EVPNet &  \textbf{96.55} & \textbf{80.89}  &  \textbf{66.71} & \textbf{4.71}  & \textbf{43.17} & \textbf{7.78} & \textbf{90.49}  \\ \cline{2-10}
		& \multirow{2}{*}{FGSM}  & \multicolumn{1}{c}{Baseline} & 95.79  &  66.53 &  5.83 &  1.94  & 56.33 & 4.94 &  90.56  \\
		& & EVPNet  &   \textbf{96.44}  & \textbf{90.80}  & \textbf{43.17} & \textbf{20.71}  & \textbf{58.22} & \textbf{18.23} & \textbf{90.78}  \\ \cline{2-10}
		& \multirow{2}{*}{PGD}   & \multicolumn{1}{c}{Baseline} & 92.47 &  74.72  & 6.25 & 1.53  &57.56 & 6.36 & 91.03  \\
		& & EVPNet & \textbf{96.06}  &   \textbf{80.45}  &   \textbf{31.30}  & \textbf{8.69} & \textbf{60.02} & \textbf{12.43} & \textbf{92.40} \\ \hline
		\multirow{6}{*}{SE-ResNet-56} & \multirow{2}{*}{Normal} & \multicolumn{1}{c}{Baseline} & 96.48 & 75.14 & 1.19 & 0.05  & 34.92 & 0.43 & 90.53  \\
		& & EVPNet  & \textbf{96.68}  &  \textbf{84.60}  & \textbf{41.94} &  \textbf{14.30} & \textbf{46.46} &\textbf{9.94} &\textbf{90.94} \\ \cline{2-10}
		& \multirow{2}{*}{FGSM}  & \multicolumn{1}{c}{Baseline} & 96.60 & 76.03 & 3.67 & 0.00 & 57.32 & 6.78 & 91.03  \\
		& & EVPNet & \textbf{97.09} & \textbf{81.53}  & \textbf{33.83} & \textbf{18.34} & \textbf{59.54} & \textbf{20.34} & \textbf{91.70} \\ \cline{2-10}
		& \multirow{2}{*}{PGD} & Baseline & 86.50 & 69.92 & 12.95 & 3.27  & 59.23 & 7.83 &  91.43 \\
		& & EVPNet & \textbf{96.39} & \textbf{77.82} & \textbf{20.96} & \textbf{7.21} & \textbf{60.34} & \textbf{13.10} & \textbf{92.06}\\ \hline
	\end{tabular}
\end{table*}

\begin{figure*}[!htbp]
	\centering
	\scriptsize
	\begin{minipage}{0.47\linewidth}
		\centering
		\scriptsize
	\subfigure[]{\label{arch-search-a5}\includegraphics[width=0.46\linewidth]{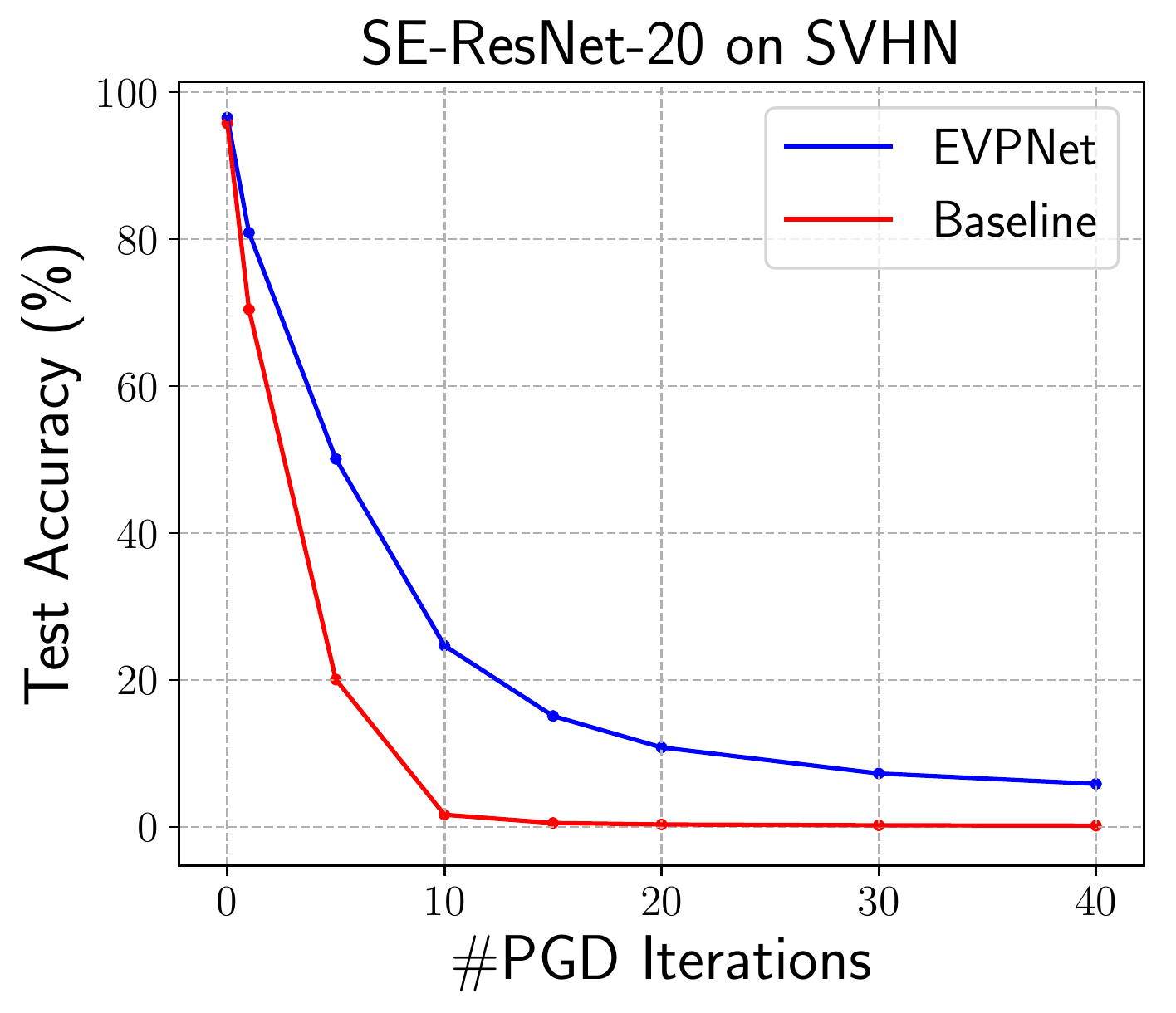}}
	\subfigure[]{\label{arch-search-a6}\includegraphics[width=0.46\linewidth]{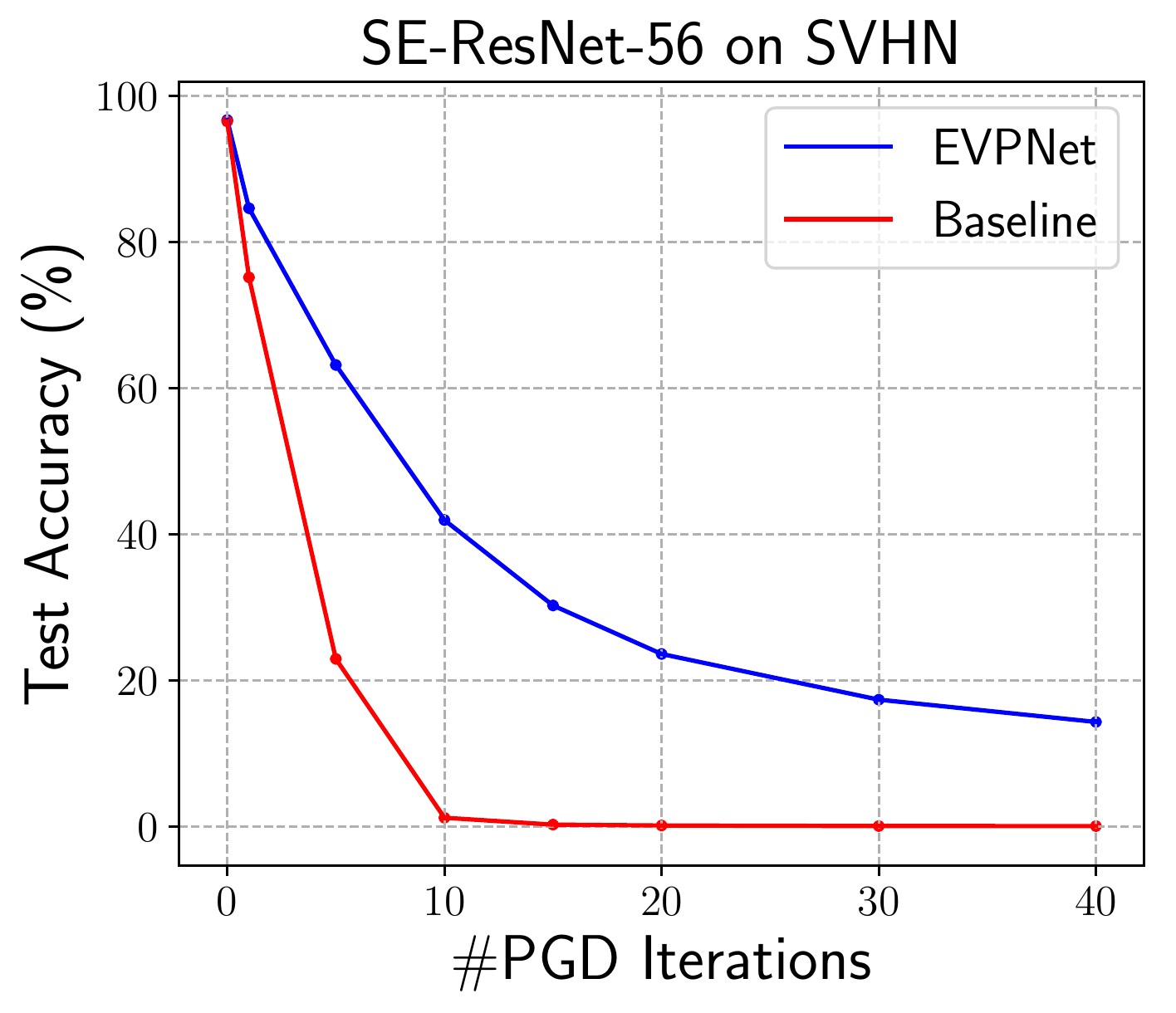}}
		\caption{Test accuracy \textit{vs} the number of PGD iterations for both networks on SVHN. Better viewed in colors.}
		\label{fig2:svhn}
	\end{minipage}
	\hspace{1ex}
	\begin{minipage}{0.47\linewidth}
		\centering
		\scriptsize
	\subfigure[]{\label{arch-search-b2}\includegraphics[width=0.46\linewidth]{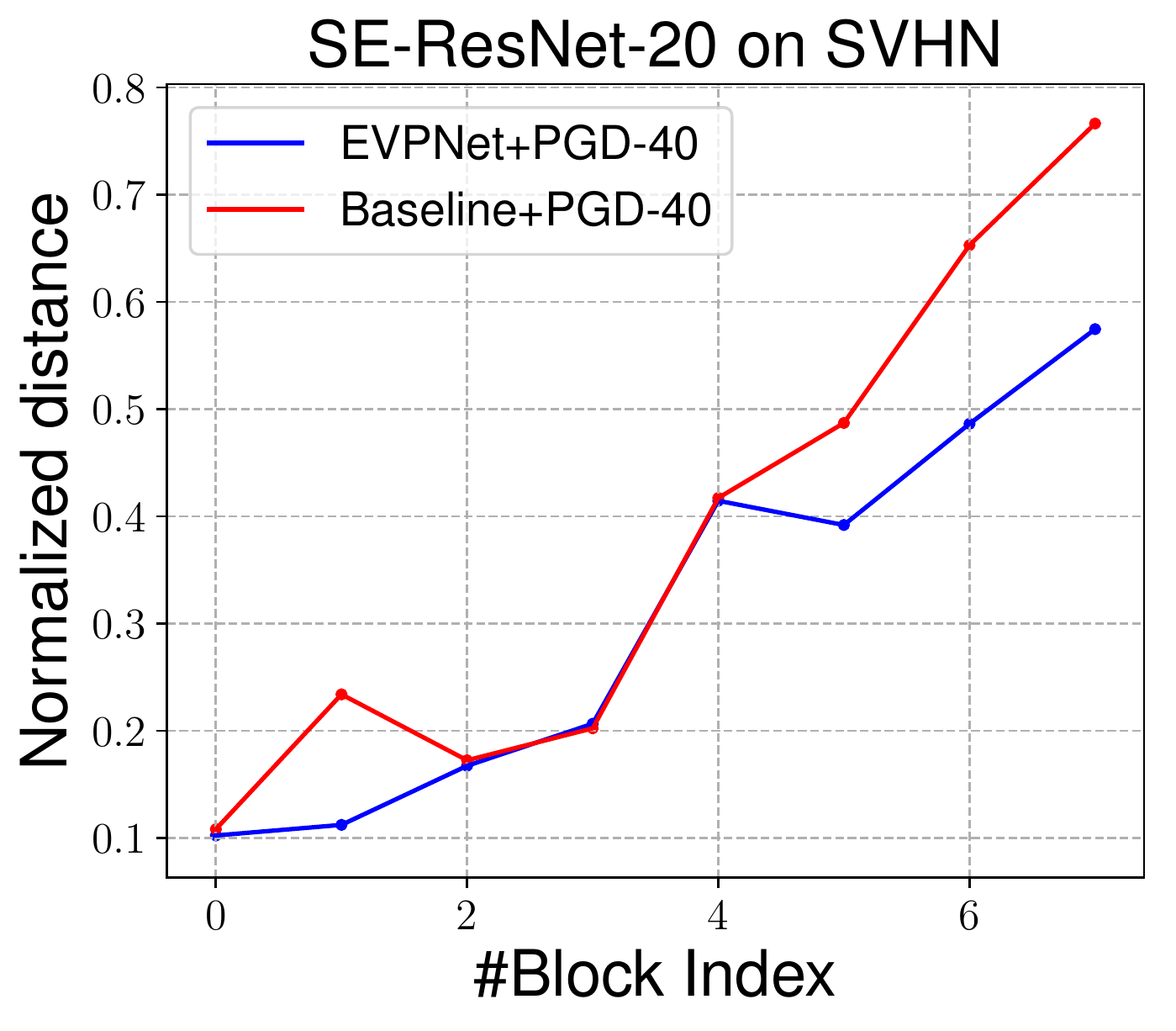}}
	\subfigure[]{\label{arch-search-b3}\includegraphics[width=0.46\linewidth]{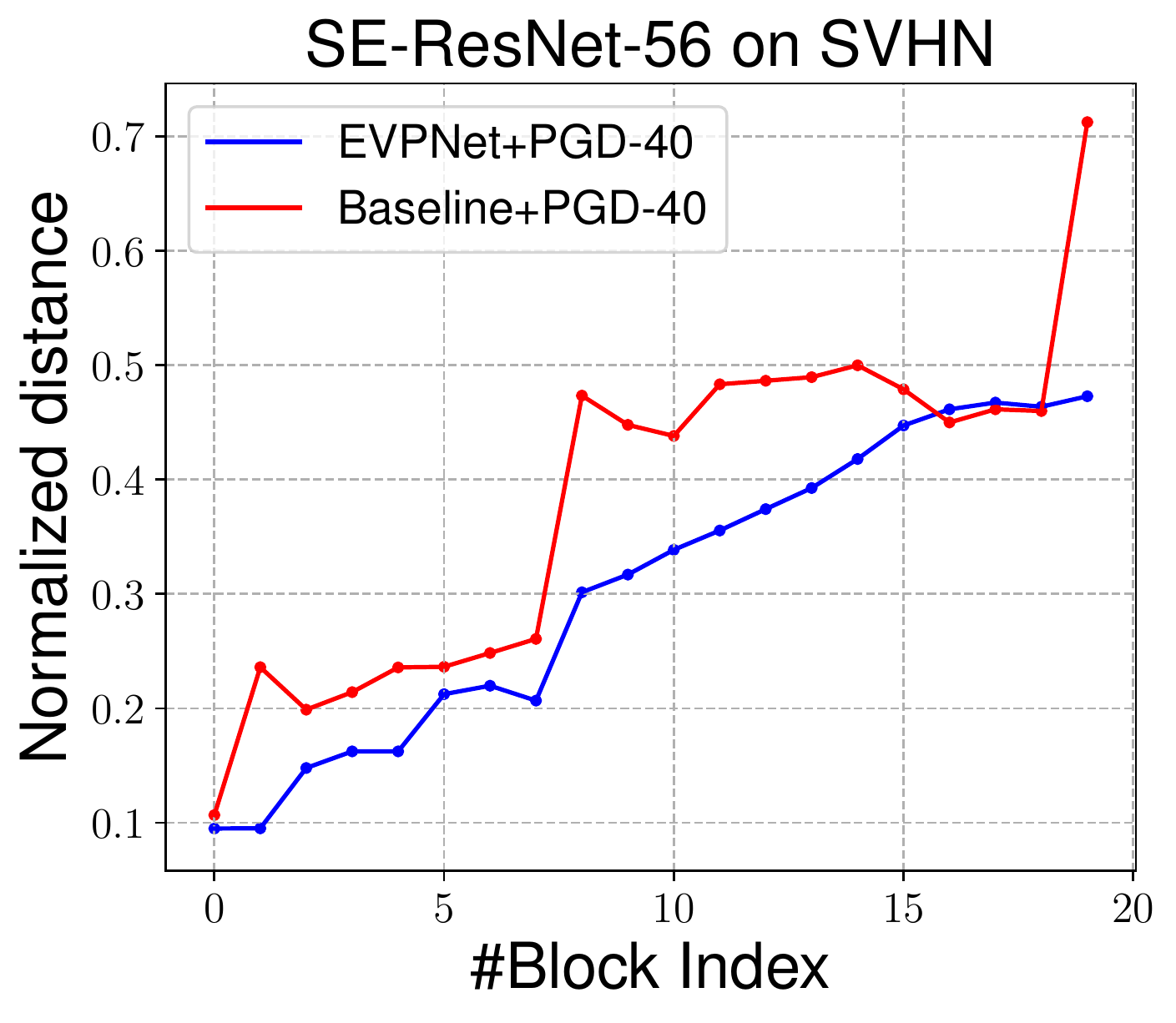}}
		\caption{
			Normalized adversarial-benign distance for feature responses between adversarial and benign examples at different res-block.}
		\label{fig3:svhn}
	\end{minipage}
	\vspace{-2ex}
\end{figure*}

We plot the test error at different PGD iterations for different evaluated networks under normal training case as shown in \autoref{fig2:svhn} for the SVHN dataset. 
\autoref{fig3:svhn} illustrates the results on the error amplification effect as \cite{liao2018guided} for SVHN.
Both figures follow the same trends as the analysis results on CIFAR-10 in the main paper. 
This justifies the generalization of our analysis for the proposed EVPNet. 

\end{appendices}

\end{document}